\documentclass[runningheads]{llncs}

\usepackage[mobile]{eccv}

\usepackage{eccvabbrv}
\usepackage{multirow}

\usepackage{graphicx}
\usepackage{booktabs}

\usepackage[accsupp]{axessibility}  

\usepackage{hyperref}
\usepackage{amsmath}
\usepackage{marvosym}
\newcommand{\corrauthor}{%
  \textsuperscript{\raisebox{0.1ex}{(\scalebox{1.2}{\Letter})}}%
}
\usepackage{tcolorbox}
\usepackage{listings}
\usepackage{amssymb}
\usepackage{orcidlink}

\begin{document}

\title{Robust Zero-shot Anomaly Detection under Limited Auxiliary Anomaly Priors} 

\titlerunning{DIVE}

\author{Guanyu Lu \and
Fang Zhou\corrauthor \and
Cheqing Jin}

\authorrunning{G. Lu et al.}

\institute{East China Normal University, Shanghai, China\\
\email{gylu@stu.ecnu.edu.cn, \{fzhou, cqjin\}@dase.ecnu.edu.cn}}

\maketitle

\begin{abstract}
  Zero-shot anomaly detection aims to identify defects in arbitrary novel domains; however, existing models assume that the auxiliary data contains a rich diversity of anomalies, neglecting the far more complex and unpredictable variations in real-world target domains. This study introduces DIVE, the first approach to investigate the scenario of limited auxiliary anomaly priors and resolve the resulting substantial performance degradation. Through a shallow-and-deep text embedding injection strategy during visual encoding, DIVE learns to abstract generic anomaly concepts shared across the auxiliary training domain and diverse target domains. Moreover, we propose a disentanglement mechanism to tackle the suboptimal alignment between visual embeddings entangled with object semantics and object-agnostic textual prompts. Experiments demonstrate that, under the setting of limited anomaly patterns in auxiliary data, DIVE outperforms SOTA baselines by up to 16.2\% and 28.5\% on two classification metrics, and 23.4\%, 24.1\%, and 47.0\% on three segmentation metrics, in terms of average performance across twelve datasets. Furthermore, it maintains highly competitive performance when auxiliary data exhibits sufficient anomaly diversity.
\end{abstract}

\section{Introduction}
\label{sec-introduction}
Anomaly detection (AD)~\cite{cao2024survey,li2025survey} plays a crucial role in identifying instances that deviate from normal patterns, forming the foundation of a wide range of applications, such as industrial defect detection~\cite{zhu2024toward,ho2024long}, medical diagnosis~\cite{cai2024rethinking,zhang2024mediclip}, and financial fraud identification~\cite{lu2024targeted}. Traditional AD approaches predominantly rely on the assumption that an adequate amount of training data can be acquired beforehand. However, compiling extensive datasets for every newly encountered scenario is often infeasible. This limitation is prominent in two scenarios: 1) accessing or centralizing training data may violate strict privacy policies, such as handling sensitive patient scans in collaborative diagnostics; 2) the target domain may intrinsically lack historical data, as seen in customized manufacturing where novel products demand immediate inspection.

\begin{figure*}[t]
\centering
\includegraphics[width=10.5cm]{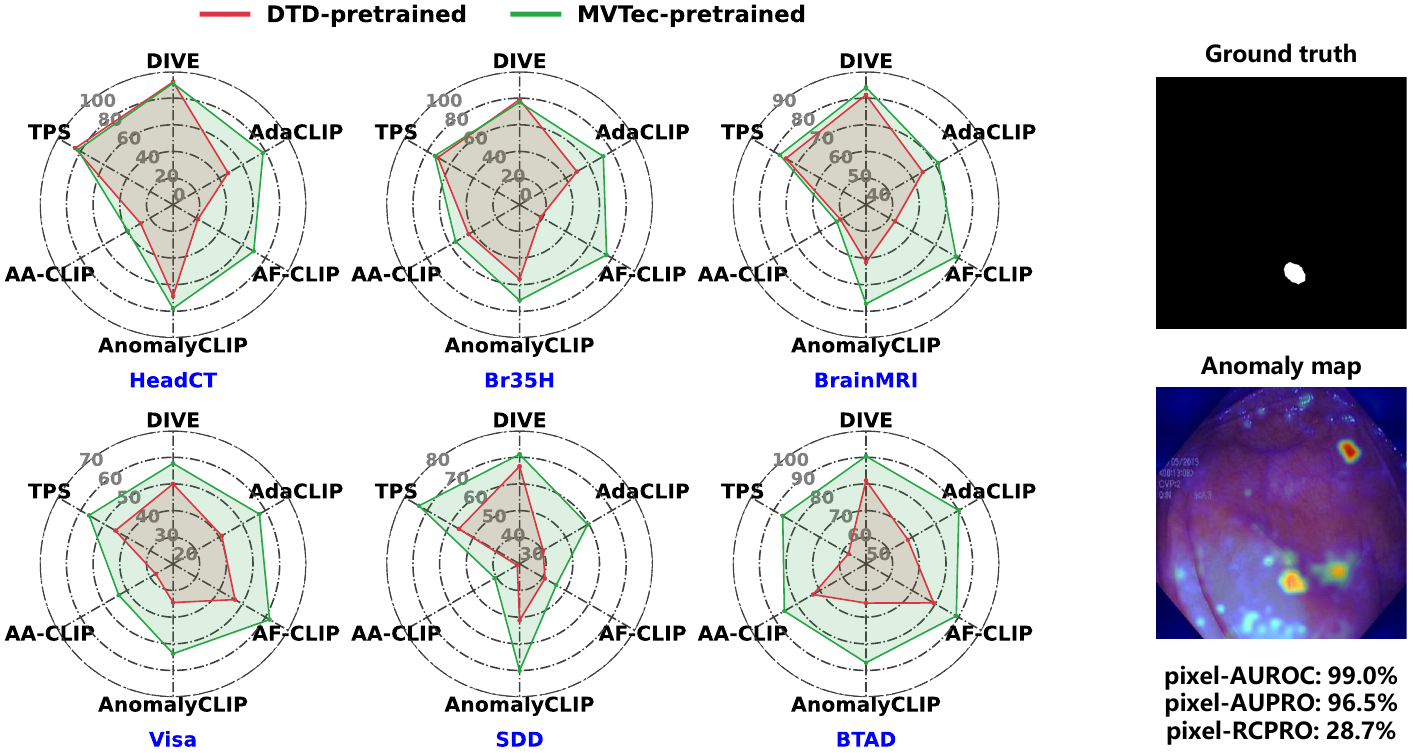}
\caption{Left: AP results of five SOTA baseline models and DIVE (our model), evaluated on six target datasets under different auxiliary pre-training settings. The green and red regions denote models pre-trained on MVTec and DTD, respectively. Right: The visualization result of AnomalyCLIP on a single test image from the ColonDB dataset, with performance measured by AUROC, AUPRO, and RCPRO (our proposed metric).}
\label{introduction}
\end{figure*}

To overcome the aforementioned limitations, Zero-shot Anomaly Detection (ZSAD) has been proposed to identify anomalies in entirely unseen target domains without any prior domain-specific data collection. Existing ZSAD methods predominantly rely on large Vision-Language Models (VLMs), such as CLIP~\cite{radford2021learning}, which are pre-trained on massive image-text pairs and provide rich open-world visual priors together with strong cross-modal alignment capabilities. Early studies primarily adopt hand-crafted prompt engineering~\cite{jeong2023winclip,chen2023zero,anovl}, aligning visual features with manually designed text prompts. This paradigm heavily depends on human expertise and task-specific prompt design, which restricts model scalability and adaptability to diverse anomaly patterns. To alleviate this issue, recent works have shifted towards learnable prompt tuning~\cite{fang2025af,zhu2025fine,zhou2023anomalyclip,ma2025aligning}, enabling the model to automatically optimize text embeddings for anomaly detection. Despite the enhanced flexibility brought by learnable prompt tuning, existing ZSAD paradigms still encounter two challenges:

\noindent
\textit{Challenge 1. Degraded generalization under limited auxiliary anomaly priors.} Existing ZSAD methods primarily leverage learnable prompts or adapter networks to fine-tune VLMs, operating under the optimistic assumption that they can fully exploit rich anomaly information in the auxiliary data.  When such auxiliary anomaly priors are scarce, these methods are highly susceptible to overfitting on specific anomaly distributions. This inevitably leads to a failure when dealing with target domains that harbor more complex defects alongside a potentially continuous influx of novel object categories. For instance, as illustrated in Fig.~\ref{introduction} (left), when the auxiliary data is switched from MVTec (comprising diverse industrial defects such as holes, stains, and cracks) to DTD (containing only texture anomalies), the average AP performance of these methods across target datasets drops by 9.9\% to 23.3\%.

\noindent
\textit{Challenge 2. Suboptimal alignment due to visual embeddings entangled with object semantics.} The feature space of pre-trained vision encoders is dominated by object semantics. When the visual embeddings are aligned with text prompts designed exclusively for anomaly evaluation, the dominant object semantics act as disruptive noise. Consequently, the model's inability to disentangle subtle defect variations from salient object identities critically hinders the learning of anomaly-discriminative representations.

To this end, we propose a novel model called \textbf{DIVE}\footnote{Code is available at https://github.com/ZhouF-ECNU/DIVE.} (\textbf{D}isentangled and text-\textbf{I}njected \textbf{V}isual \textbf{E}mbeddings) for ZSAD. To overcome the generalization bottleneck caused by insufficient abnormality coverage (Challenge 1), DIVE introduces a shallow-and-deep text embedding injection strategy. At the shallow level, the cross-modal projection of learnable semantic text prompts into visual prompts enables synergistic optimization for effective generalization across unseen object semantics. Subsequently, by selectively fusing each patch token extracted at deep layers with the text embeddings of LLM-generated descriptions about normality and abnormality through a cross-attention formulation, the model is empowered to abstract generic anomaly concepts from specific auxiliary visual patterns. For instance, a texture anomaly might share similarities in shape and contour with an industrial fabric stain or a medical tumor shadow; thus, its corresponding patch can match with enriched text prompts like \texttt{"a photo of an object with a stain"} or \texttt{"a photo of an object with a shadow"}. As verified in Fig.~\ref{introduction} (left), benefiting from this strategy, DIVE restricts the average performance penalty to a mere 5.3\%, effectively reducing the degradation magnitude by approximately 46.5\% to 77.3\% compared to the drastic drop observed in baselines under the DTD-pretrained setting. Furthermore, to eliminate the interference of object semantics on AD tasks (Challenge 2), a disentanglement mechanism is applied to decouple the global visual embeddings into independent anomaly-state and object-semantic subspaces, where each is optimized for cross-modal alignment under the guidance of its respective text prompt learner. By imposing orthogonal constraints, the state branch focuses exclusively on visual cues essential for anomaly identification.

This paper additionally introduce a segmentation evaluation metric to resolve the flaws of the standard AUPRO metric~\cite{bergmann2019mvtec,bergmann2020uninformed} widely adopted in prior works~\cite{fang2025af,zhou2023anomalyclip,zhu2025fine,gong2025fe,ma2025aligning,he2025rareclip}, which ignores false positives in normal regions and over segmentation of anomaly regions. As depicted in Fig.~\ref{introduction} (right), while AnomalyCLIP highlights three primary anomalous regions on a given test instance, two are stark false positives (including the region with the highest anomaly score).  Since the AUPRO metric solely focuses on the coverage of ground-truth anomalous regions, it yields a misleadingly inflated score of 96.5\%. By integrating a region-wise prediction quality penalty, the proposed metric RCPRO circumvents this issue, restricting the score of the aforementioned test case to a realistic 28.7\%.

In summary, this paper makes the following key contributions:
\begin{itemize}
\item To the best of our knowledge, this work pioneers the study of ZSAD under the setting of limited auxiliary anomaly priors and analyzes the severe generalization bottlenecks induced by this constraint. This realistic scenario reflects the unpredictable variations and the continual emergence of novel anomaly patterns in real-world target domains.

\item We propose DIVE, a tailored model for this challenging scenario, with two core components for visual embedding learning: text embedding injection to enable generalization to unseen object categories and abstraction of generic anomaly concepts, and a disentanglement mechanism to alleviate semantic interference in global embeddings.

\item We formulate RCPRO, a novel metric that rectifies AUPRO's leniency towards false positives in normal regions and over-segmentation of anomalous regions. Experiments on twelve datasets verify the superiority and robustness of DIVE under both limited and sufficient anomaly diversity settings.
\end{itemize}

\section{Related Work}
\subsection{Anomaly Detection}
Conventional AD are typically built on the assumption of \textit{i.i.d.} training and testing data, meaning they are sampled from the same domain. Based on the availability of labeled anomalies, these methods can be categorized into unsupervised~\cite{xu2023fascinating,xu2023deep,zhou2024msflow,guo2025dinomaly} and semi-supervised~\cite{lu2024robust,wei2024gad,shou2025read,jiang2023anomaly} settings. Such a closed-set paradigm leads to limited generalization across diverse domains, necessitating costly retraining when deployed in new scenarios.

To achieve broader applicability, Zero-shot Anomaly Detection has emerged as a promising direction. Generally, current ZSAD methods can be categorized into two paradigms based on their utilization of Large Language Models (LLMs): (i) prompting-based detection~\cite{zhang2024gpt,gu2024anomalygpt}, which directly queries LLMs to generate descriptive responses indicating anomalies, yet suffers from the imprecise articulation of fine-grained anomaly patterns; (ii) contrasting-based detection~\cite{zhu2024toward,li2024promptad} offers a more effective alternative by leveraging pre-trained VLMs (e.g., CLIP) to compute similarity between image features and textual prompt embeddings for anomaly identification. As the pioneering work applying CLIP to ZSAD, WinCLIP~\cite{jeong2023winclip} aligns window-level visual features with hand-crafted text prompts, whose manual construction restricts adaptability and generalization, a limitation also shared by VAND~\cite{chen2023zero} and AnoVL~\cite{anovl}. To solve this issue, AnomalyCLIP~\cite{zhou2023anomalyclip} introduces an object-agnostic prompt design, where specific portions of the text prompts are substituted with learnable tokens for dynamic optimization. Furthermore, some recent works like TPS~\cite{ma2025aligning} and FAPrompt~\cite{zhu2025fine} focus on designing fine-grained text prompts to capture nuanced abnormal patterns and enhance local grounding. While these approaches have somewhat contributed to improvements in ZSAD, their generalization heavily relies on the assumption of sufficient anomaly diversity in the auxiliary data, causing severe performance degradation when target domains exhibit significant shifts.

\subsection{Prompt Engineering}
Vision-Language Models have exhibited impressive zero-shot capabilities, which are largely contingent upon effective prompt engineering. Early attempts relied on manually designed static text templates~\cite{radford2021learning}, which are not only labor-intensive but also highly sensitive to specific wording choices. To overcome the limitations of manual design, CoOp~\cite{zhou2022learning} pioneered the use of continuous learnable vectors in the text encoder. Subsequent research further advanced textual prompt tuning by learning prompt distributions (ProDA~\cite{lu2022prompt}), incorporating knowledge guidance (KgCoOp~\cite{yao2023visual}), or leveraging optimal transport (PLOT~\cite{chen2022plot}). In response to the limited generalization of static learned prompts to unseen classes, CoCoOp~\cite{zhou2022conditional} introduced instance-conditional dynamic prompting, where learnable tokens are conditioned on individual image features. Recognizing that single-modality adaptation is suboptimal for complex cross-modal alignment, MaPLe~\cite{khattak2023maple} explores multi-modal prompt learning, jointly optimizing learnable prompts in both vision and language branches. In this paper, we construct visual embeddings through text-injected synergistic learning and disentanglement, making CLIP more suitable for downstream AD tasks and enabling it to abstract anomaly concepts that are common across diverse domains.

\section{Problem Formulation}
Consider an auxiliary training dataset $\mathcal{D}_{train} = \{(x_i, y_i, \mathbf{G}_i)\}_{i=1}^{l}$ comprising $l$ samples from a source domain, where $x_i \in \mathbb{R}^{H \times W \times 3}$ is the input image, $y_i \in \{0, 1\}$ denotes the image-level label (with 0 representing normal and 1 representing anomalous), and $\mathbf{G}_i \in \{0, 1\}^{H \times W}$ represents the ground-truth pixel-level anomaly mask. The objective of Zero-Shot AD is to learn a generalized model on $\mathcal{D}_{train}$ capable of handling multiple test datasets $\{\mathcal{D}_{test}^k\}_{k=1}^K$ derived from $K$ distinct unseen target domains that are disjoint from the source domain. During inference, for any query image $x \in \mathcal{D}_{test}$, the model is expected to simultaneously predict a scalar anomaly score $s \in [0, 1]$ indicating the probability of the image being anomalous, and a pixel-wise anomaly map $\mathbf{M} \in [0, 1]^{H \times W}$ that localizes the anomalous regions. For both the image-level score $s$ and the pixel values in $\mathbf{M}$, a higher value indicates a higher degree of abnormality.

\section{Methodology}
\begin{figure*}[!t]
\centering
\includegraphics[width=12.2cm]{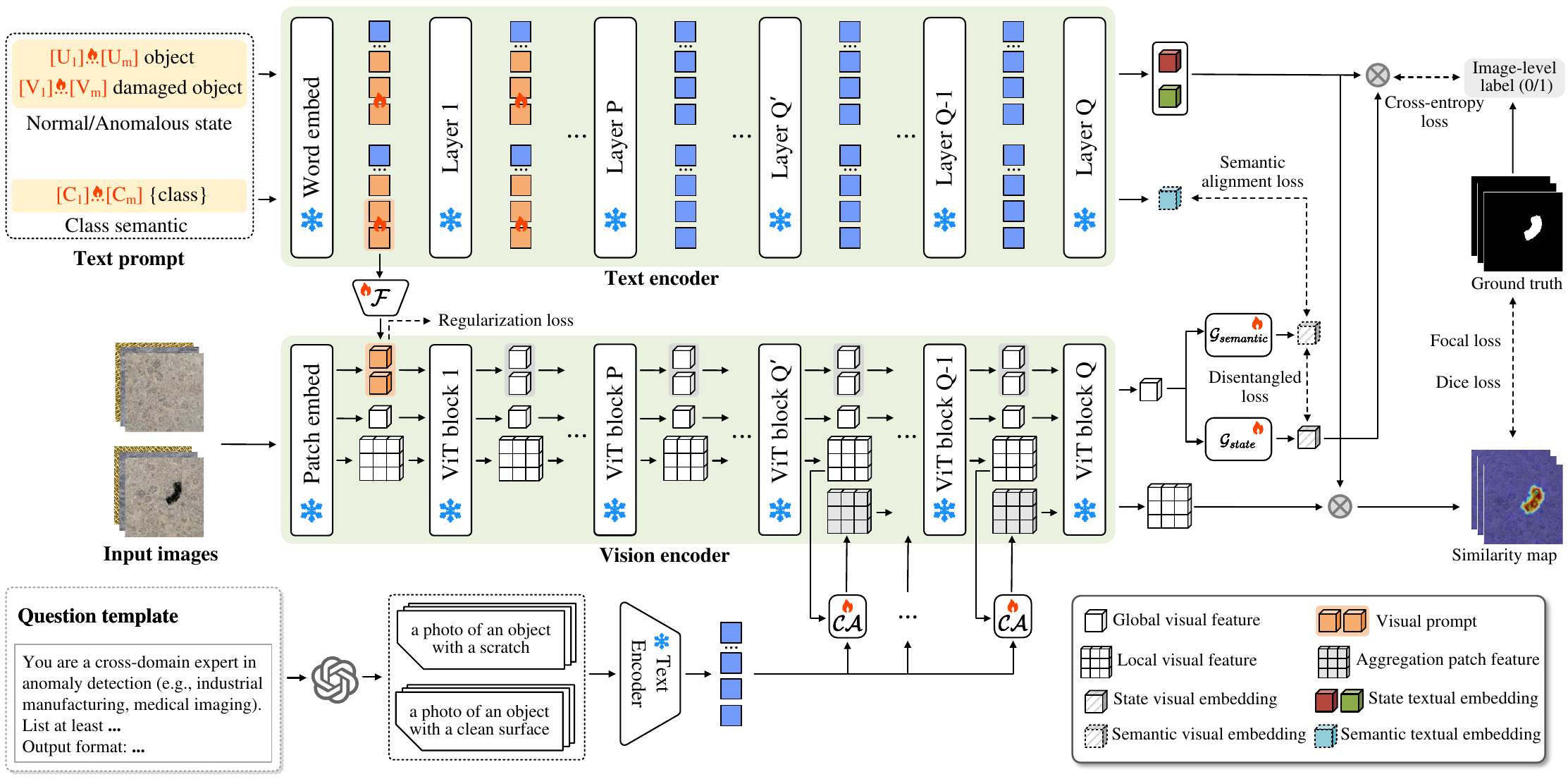}
\caption{The workflow of DIVE.}
\label{workflow}
\end{figure*}

The overall workflow of DIVE is illustrated in Fig.~\ref{workflow}. Following the CLIP-based AD paradigm~\cite{zhou2023anomalyclip,cao2024adaclip,ma2025aligning,ma2025aa,zhu2025fine}, DIVE aligns global and local visual embeddings with designated ``normal'' and ``abnormal'' text prompts. Notably, a text embedding injection strategy (Vision-Language (V-L) coupling function $\mathcal{F}$ and cross-attention networks $\mathcal{CA}$ in Fig.~\ref{workflow}) is applied to enhance the model's generalization to unseen object categories and complex anomaly patterns. Furthermore, to prevent visual embeddings from entangling with object semantics, we introduce a disentanglement mechanism (branch networks $\mathcal{G}_{state}$ and $\mathcal{G}_{semantic}$ in Fig.~\ref{workflow}) to capture global visual information related to anomaly identification.

\subsection{Independent Parallel Text Prompting}
In traditional VLMs, text prompt templates typically entangle anomaly descriptions with specific object identities (e.g., \texttt{"a photo of a damaged \{class\}"}) \cite{jeong2023winclip}. To decouple these semantics, AnomalyCLIP~\cite{zhou2023anomalyclip} discards the token \texttt{\{class\}} from the prompts. Alternatively, it introduces a set of learnable context vectors to construct two object-agnostic text prompt templates:
\begin{equation*}
    t_n = [U_1][U_2]\ldots[U_m][object], \quad t_a = [V_1][V_2]\ldots[V_m][damaged][object].
\end{equation*}
However, directly aligning visual embeddings extracted by a pre-trained vision encoder with object-agnostic text prompts is suboptimal, as it forces semantic-dominated visual embeddings into a textual space exclusively designed for anomaly detection. To address this issue, we further design an independent semantic-aware prompt learner on the text side. Operating in parallel with the aforementioned state-aware prompt learner, this module is specifically dedicated to capturing the semantic category information of target objects. This parallel design provides textual anchors for the subsequent precise disentanglement of state-related visual embeddings and semantic-related visual embeddings at the visual end. Specifically, for an auxiliary training data comprising $R$ distinct object categories, we formulate the semantic-aware text prompt template using a separate set of learnable context vectors as follows:
\begin{equation*}
    t_c = [C_1][C_2]\ldots[C_m][\text{class}_r], \quad r \in \{1, 2, \ldots, R\},
\end{equation*}
where $[\text{class}_r]$ represents the word token for the $r$-th specific object class.

To encourage the textual context prompts in two prompt learner to better generalize to both anomaly detection and object recognition tasks, a new set of learnable tokens is introduced into each transformer block $\texttt{T}_j$ of the text encoder up to depth $P$:
\begin{equation}
\scalebox{0.78}{$
    \begin{gathered} 
        [\_, U_{j}^i|_{h+1}^m, N_j] = \texttt{T}_j([U_{j-1}^i|_{1}^h, U_{j-1}^i|_{h+1}^m, N_{j-1}]), \quad [\_, V_{j}^i|_{h+1}^m, A_j] = \texttt{T}_j([V_{j-1}^i|_{1}^h, V_{j-1}^i|_{h+1}^m, A_{j-1}]), \\
        [\_, C_{j}^i|_{h+1}^m, S_j] = \texttt{T}_j([C_{j-1}^i|_{1}^h, C_{j-1}^i|_{h+1}^m, S_{j-1}]), \quad j=1,2,\dots,P,
    \end{gathered}
$}
\end{equation}
where $N$, $A$, and $S$ denote the fixed input tokens; $U^i|_{1}^h$, $V^i|_{1}^h$, and $C^i|_{1}^h$ represent the newly introduced learnable tokens with a sequence length of $h$; and $[\cdot ,\cdot ,\cdot ]$ denotes the concatenation operation. Beyond the $P^{\text{th}}$ Transformer layer, each subsequent layer takes the prompts from its preceding layer as input:
\begin{equation}
\scalebox{0.9}{$
\begin{gathered}
    [U_{j'}, N_{j'}] = \texttt{T}_{j'}([U_{{j'}-1}, N_{{j'}-1}]), \quad [V_{j'}, A_{j'}] = \texttt{T}_{j'}([V_{{j'}-1}, A_{{j'}-1}]), \\
    [C_{j'}, S_{j'}] = \texttt{T}_{j'}([C_{{j'}-1}, S_{{j'}-1}]), \quad j'=P+1,...,Q.
\end{gathered}
$}
\end{equation}

Ultimately, by feeding the final tokens $N^m_{Q}, A^m_{Q}$, and $C^m_{Q}$ into the text projection head, the two parallel prompt learners generate the state textual embeddings $z_n, z_a$ and the semantic textual embedding $z_c$, respectively.

\subsection{Textual Embedding Injection}
Lacking prior knowledge regarding unseen categories and normal/abnormal patterns in the target domain, the pre-trained visual encoder suffers from restricted generalization capabilities in novel scenarios, which incurs a substantial performance degradation. Inspired by the multi-modal prompting mechanism of MaPLe~\cite{khattak2023maple}, learnable text prompting are introduced to guide visual feature extraction, which preserves the cross-modal alignment capability for novel object categories. Given that the shallow layers of Vision Transformer (ViT) are primarily responsible for capturing coarse-grained local structures such as edges and textures, we map the first $h$ learnable tokens of the semantic text prompt into visual prompts via a V-L coupling function $\mathcal{F}$ before the first layer, injecting them into the visual feature extraction process, i.e., $[c_{1}, E_1, \tilde{C}_1^{i}|_1^h] = \texttt{ViT}_1([c_{0}, E_0, \mathcal{F}(C_0^i|_1^h)])$, 
where $c$ denotes the global visual feature, and $E$ represents the local visual feature comprising $H \times W$ patch tokens.

While the shallow injection of semantic text prompts improves generalization to unseen object categories, the model remains highly susceptible to overfitting the known anomaly patterns in the auxiliary dataset. When encountering novel variants, this overfitting directly leads to the misclassification of unseen anomalies and false alarms on normal samples. Given that the deep layers of ViT possess strong conceptual abstraction capabilities to represent object-agnostic state features, we directly fuse the rich LLM-generated typological descriptions of normality and abnormality with the patch tokens via cross-attention in these deep layers. This process compels the model to abstract the concrete visual patterns from the auxiliary training data and map them onto more generalizable, universal anomaly concepts. Formally, executing this cross-modal fusion after the $Q'^{\text{th}}$ ViT block:
\begin{equation}
\scalebox{0.77}{$
    [c_{d}, E_d, \tilde{C}_d] = \texttt{ViT}_d([c_{d-1}, E_{d-1}, \tilde{C}_{d-1}]), \quad d=2, 3, \dots, Q',
$}
\end{equation}

\begin{equation}
\scalebox{0.76}{$
    [c_{d'}, E_{d'}, \tilde{C}_{d'}] = \texttt{ViT}_{d'}([ c_{d'-1}, E_{d'-1} + \gamma \cdot \mathcal{CA}_{d'-1}(E_{d'-1}, Z_{g} W_{proj}^\top, Z_{g} W_{proj}^\top), \tilde{C} _{d'-1}]), \quad d'=Q'+1, \dots, Q,
$}
\end{equation}
where $Z_{g}$ denotes the text embeddings obtained by feeding diverse LLM-generated descriptions of normality and abnormality (the question template are detailed in Appendix~\ref{Appendix-A.1}) into a pre-trained CLIP text encoder, and $W_{proj}^\top$ is a learnable linear projection matrix that maps these textual embeddings into the visual feature space; $\mathcal{CA}(\cdot,\cdot,\cdot)$ denotes the cross-attention operation, utilizing the visual patch tokens as the query, and the projected text features as the key and value. Notably, these cross-modal state textual priors are injected via a residual connection, where $\gamma$ is a hyperparameter designed to balance the original visual features and the injected textual information.

To extract the final visual embeddings, the global visual feature $z_x$ and local patch features $z_e^{(i,j)} (i \in [1,H], j \in [1,W])$  are derived by processing the output tokens $c_{Q}$ and $E_{Q}$ via the image projection head.

\subsection{Visual Embedding Disentanglement}
The global visual feature $z_x$ extracted by the vision encoder is a highly entangled embedding, inherently coupling the object identity semantics with the anomaly state information. Directly aligning this entangled embedding with text prompts would inevitably blur the boundary between normal and anomalous instances across different categories. Therefore, we design two parallel learnable Multi-Layer Perceptron (MLP) networks, denoted as $\mathcal{G}_{state}(\cdot)$ and $\mathcal{G}_{semantic}(\cdot)$, to explicitly disentangle $z_x$ into two independent subspaces. This process yields the state visual embedding $z_{st}$ and the semantic visual embedding $z_{se}$, i.e.,
\begin{equation}
\scalebox{0.98}{$
z_{st} = z_x + \Delta z_{st} = z_x + \mathcal{G}_{state}(z_x), \quad z_{se} = z_x + \Delta z_{se} = z_x + \mathcal{G}_{semantic}(z_x).
$}
\end{equation}
Here, instead of learning an absolute transformation, we formulate the disentanglement through a residual learning paradigm, where $\Delta z_{st}$ and $\Delta z_{se}$ represent the learned residual offsets for the state and semantic subspaces, respectively. This not only mitigates the risk of catastrophic forgetting of pre-trained visual concepts but also stabilizes the optimization process.

\subsection{Training and Inference}
\label{Training and Inference}
During the training phase, the backbone parameters of the pre-trained vision and text encoders are strictly frozen to retain their open-world generalization capabilities. Only the state-aware and semantic-aware prompt learners, the V-L coupling function $\mathcal{F}$, the introduced cross-attention layers, and the visual MLP projection heads ($\mathcal{G}_{state}$ and $\mathcal{G}_{semantic}$) are trainable.

For the anomaly classification task, we compute the cosine similarity between the state visual embedding $z_{st}$ and the normal/abnormal state textual embeddings $\{z_n, z_a\}$. The state prediction probability $p_{st}$ is formulated as: $p_{st} = \text{softmax}\left( [z_{st}^\top z_n, z_{st}^\top z_a] / \tau \right)$, where $\tau$ is a temperature hyperparameter. The state-aware alignment is optimized using the cross-entropy loss: $\mathcal{L}_{state} = \text{CE}(p_{st}, y_i)$.

Similarly, to ensure that the semantic branch accurately captures the object identity, we align the semantic visual embedding $z_{se}$ with the semantic textual embeddings $\{z_c^r\}_{r=1}^R$ via a contrastive learning objective. Specifically, $z_{se}$ is pulled closer to the textual embedding of its corresponding ground-truth category $y_{cls}$, while being pushed apart from those of other mismatched categories:
$\mathcal{L}_{sem} = -\log \frac{\exp(z_{se}^\top z_c^{y_{cls}} / \tau)}{\sum_{r=1}^R \exp(z_{se}^\top z_c^r / \tau)}$,
where $z_c^{y_{cls}}$ represents the positive textual embedding anchor associated with the actual class of the input image.

To enforce the disentanglement between the object semantics and anomaly states, we apply an orthogonal constraint on the learned residual offsets. The disentanglement loss is calculated using the squared cosine similarity between the two offset vectors: 
$\mathcal{L}_{dis} = \left( \frac{\Delta z_{st}^\top \Delta z_{se}}{\left\| \Delta z_{st} \right\|_2 \left\| \Delta z_{se} \right\|_2} \right)^2$.

For the pixel-level segmentation task, we compute the similarity map $\mathbf{S}$ by calculating the similarity scores between the local patch features $z_e^{(i,j)}$ and the state textual embeddings $\{z_n, z_a\}$. Focal loss~\cite{lin2017focal} and Dice loss~\cite{li2020dice} are jointly applied to optimize the anomaly segmentation map against the ground-truth mask $\mathbf{G}$: $\mathcal{L}_{seg} = \text{Focal}(Up(\mathbf{S}), \mathbf{G}) + \text{Dice}(Up(\mathbf{S}_{a}), \mathbf{G}) + \text{Dice}(Up(\mathbf{S}_{n}), \mathbf{I} - \mathbf{G})$,
where $\mathbf{S}_{n}$ and $\mathbf{S}_{a}$ denote the normal and abnormal probability channels of $\mathbf{S}$, respectively, and $\mathbf{I}$ is the full-one matrix. The operator $Up(\cdot)$ represents the upsampling operation used to restore the feature resolution to the original image size. Furthermore, to prevent the visual prompt tokens from overpowering the original visual features and causing instability, we impose a regularization constraint $\mathcal{L}_{reg} = \frac{1}{|\tilde{C}|} ||\tilde{C}||_2^2$ on the generated visual prompts $\tilde{C}$, where $|\tilde{C}|$ denotes the total number of elements in the prompt tensor.

The overall objective function is optimized as follows:
$\mathcal{L} = \mathcal{L}_{state} + \lambda\mathcal{L}_{sem} + \lambda^{\prime}\mathcal{L}_{dis} + \mathcal{L}_{seg} + \lambda^{\prime\prime}\mathcal{L}_{reg}$,
where $\lambda$, $\lambda^{\prime}$ and $\lambda^{\prime\prime}$ are trade-off weighting coefficients.

In the detection phase, inference is performed using the frozen model weights optimized on the auxiliary dataset. Given a query test image $x$, we first calculate its image-level anomaly score $s \in [0, 1]$. We directly apply the abnormal channel of the state prediction probability as the anomaly score, which tends toward 1 when the abnormal state textual embedding $z_a$ is highly aligned with the state visual embedding $z_{st}$:
\begin{equation}
    s = \frac{\exp(z_{st}^\top z_a / \tau)}{\exp(z_{st}^\top z_n / \tau) + \exp(z_{st}^\top z_a / \tau)}.
\end{equation}
For pixel-wise predictions, the final-layer patch features $z_e^{(i,j)}$ are used to compute a similarity map. Its abnormal channel $\mathbf{S}_{a}$ is then upsampled and spatially smoothed to generate the final anomaly map: $\mathbf{M} = {G}_{\sigma} \left( Up(\mathbf{S}_{a}) \right)$, where ${G}_{\sigma}$ represents a Gaussian filter with a standard deviation $\sigma$ designed to smooth the final segmentation boundaries and eliminate noise.

\section{Experiments}
\subsection{Experimental Setup}
\subsubsection{Datasets and Baselines.} To comprehensively evaluate the proposed DIVE, we selected twelve publicly available datasets, spanning both industrial defect detection (MVTec~\cite{bergmann2019mvtec}, DTD~\cite{aota2023zero}, Visa~\cite{zou2022spot}, MPDD~\cite{jezek2021deep}, SDD~\cite{tabernik2020segmentation}, and BTAD~\cite{Mishra2021VTADLAV}) and clinical medical imaging (HeadCT~\cite{salehi2021multiresolution}, Br35H~\cite{hamada2020br35h}, BrainMRI~\cite{salehi2021multiresolution}, ISIC~\cite{gutman2016skin}, ColonDB~\cite{tajbakhsh2015automated}, and TN3K~\cite{gong2021multi}). The proposed DIVE is compared against several state-of-the-art baselines representing the latest advancements in ZSAD: AnomalyCLIP~\cite{zhou2023anomalyclip}, AdaCLIP~\cite{cao2024adaclip}, AF-CLIP~\cite{fang2025af}, AA-CLIP~\cite{ma2025aa}, and TPS~\cite{ma2025aligning}. Recognizing that anomalous events are scarce in real-world applications, these datasets were resampled to simulate an imbalanced setting, where anomalies account for only a minor proportion of the entire distribution. The exceptions are ISIC, ColonDB, and TN3K. Given that they are exclusively utilized for segmentation evaluation and contain no normal instances, we follow their original testing setups. Further statistical details for the datasets, along with descriptions of the baseline models, are provided in Appendix~\ref{Appendix-A.2}.

\subsubsection{Evaluation Metrics.} Following standard evaluation procedures in ZSAD, we assess performance at both the image and pixel levels. For image-level anomaly classification, the Area Under the Receiver Operating Characteristic curve (AUROC) and Average Precision (AP) are reported. For pixel-level anomaly segmentation, AUROC along with the Area Under the Per-Region Overlap (AUPRO)~\cite{bergmann2019mvtec,bergmann2020uninformed} is used to evaluate localization accuracy.

We introduce a new segmentation metric, Region-Calibrated PRO (RCPRO), designed to overcome key limitations of the AUPRO metric (as analyzed in Section~\ref{sec-introduction}). First, a series of binary anomaly maps is generated by varying the threshold from the minimum to the maximum anomaly score. For each threshold, two core curve coordinates are then computed: the X-axis represents the standard PRO score, and the Y-axis represents the Region-Calibrated score, which quantifies the average prediction quality across all predicted anomalous regions. Specifically, if the maximum overlap between a predicted region and any ground-truth anomaly falls below a tolerance threshold (set to 0.05 of the matched ground-truth area in our setting), the predicted region receives a score of zero, thereby penalizing false positives in normal regions. On the other hand, if a correctly located prediction expands excessively beyond the ground-truth boundary, its score decays inversely proportional to the excess area ratio to penalize over-segmentation. Ultimately, the RCPRO value is defined as the area under the curve formed by these coordinates. Detailed formulas are provided in Appendix~\ref{Appendix-A.3}.

\begin{table}[t]
\caption{Performance comparison of DIVE with baseline models utilizing DTD as auxiliary data for pre-training, evaluated across two tasks: image-level classification (AUROC, AP) and pixel-level segmentation (AUROC, AUPRO, RCPRO). The best results are highlighted in \textbf{bold} and the second-best results are \underline{underlined}.}
\label{table1}
\scalebox{0.63}{
\begin{tabular}{c|c|cccccc}
\hline
Task                                                                                     & Dataset  & \begin{tabular}[c]{@{}c@{}}AnomalyCLIP\\ (ICLR' 2024)\end{tabular} & \begin{tabular}[c]{@{}c@{}}AdaCLIP\\ (ECCV' 2024)\end{tabular} & \begin{tabular}[c]{@{}c@{}}AF-CLIP\\ (MM' 2025)\end{tabular} & \begin{tabular}[c]{@{}c@{}}AA-CLIP\\ (CVPR' 2025)\end{tabular} & \begin{tabular}[c]{@{}c@{}}TPS\\ (AAAI' 2025)\end{tabular} & DIVE                   \\ \hline
\multirow{9}{*}{\begin{tabular}[c]{@{}c@{}}Classification\\ (image-level)\end{tabular}}  & HeadCT   & (95.1,   68.6)                                                  & (84.1, 47.7)                                                  & (70.9, 21.6)                                                & (75.4, 28.1)                                                  & (\underline{97.9}, \underline{84.8})                      & (\textbf{99.1}, \textbf{92.1})         \\
                                                                                         & Br35H    & (91.6, 55.8)                                                    & (92.2, 49.7)                                                  & (59.6, 18.4)                                                & (80.5, 44.2)                                                  & (\underline{94.8}, \underline{70.8})                      & (\textbf{97.3}, \textbf{78.9})         \\
                                                                                         & BrainMRI & (89.9, 61.6)                                                    & (92.8, 64.6)                                                  & (81.8, 52.5)                                                & (83.1, 51.1)                                                  & (\underline{93.6}, \underline{74.7})                      & (\textbf{96.4}, \textbf{81.2})         \\
                                                                                         & MVTec    & (81.7, 55.1)                                                    & (84.8, 63.9)                                                  & (\underline{88.2}, \underline{70.0})                        & (78.4, 56.1)                                                  & (85.5, 69.5)                                              & (\textbf{89.3}, \textbf{73.9})         \\
                                                                                         & Visa     & (66.8, 34.5)                                                    & (71.2, 41.1)                                                  & (\underline{76.3}, \underline{46.8})                        & (60.8, 27.5)                                                  & (75.0, 45.0)                                              & (\textbf{77.8}, \textbf{50.0})         \\
                                                                                         & MPDD     & (58.0, 21.9)                                                    & (63.0, 21.2)                                                  & (\underline{68.2}, 21.2)                                    & (43.5, 15.3)                                                  & (65.6, \underline{25.0})                                  & (\textbf{71.8}, \textbf{30.1})         \\
                                                                                         & SDD      & (\underline{79.1}, 51.2)                                        & (72.0, 40.2)                                                  & (75.2, 41.0)                                                & (72.1, 31.0)                                                  & (78.8, \underline{56.3})                                  & (\textbf{81.7}, \textbf{66.6})         \\
                                                                                         & BTAD     & (72.0, 64.7)                                                    & (80.4, 68.1)                                                  & (\textbf{86.2}, \underline{79.3})                           & (74.8, 73.0)                                                  & (61.6, 57.6)                                              & (\underline{85.2}, \textbf{81.3})      \\ \cline{2-8} 
                                                                                         & Average  & (79.3,   51.7)                                                  & (80.1, 49.6)                                                  & (75.8, 43.9)                                                & (71.1, 40.8)                                                  & (\underline{81.6}, \underline{60.5})                      & (\textbf{87.3}, \textbf{69.3})         \\ \hline
\multirow{9}{*}{\begin{tabular}[c]{@{}c@{}}Segmentation\\ (pixel-level)\end{tabular}}    & ISIC     & (70.7,   45.5, 41.2)                                            & (\underline{83.5}, 15.5,   66.5)                              & (74.1, 64.0,   64.1)                                        & (80.2, \underline{71.8}, \textbf{89.3})                       & (70.2, 45.2,   61.8)                                      & (\textbf{86.4}, \textbf{76.5},   \underline{74.2}) \\
                                                                                         & ColonDB  & (62.9, 26.2,   15.5)                                            & (76.3, 17.3, 30.1)                                            & (72.7, 58.8, 44.4)                                          & (\underline{78.7}, \underline{63.2}, \textbf{59.6})           & (57.8, 18.6, 26.2)                                        & (\textbf{81.8}, \textbf{68.2}, \underline{56.2})   \\
                                                                                         & TN3K     & (69.6, 21.5,   21.1)                                            & (72.6, 8.1, 54.0)                                             & (65.6, 44.9, 36.4)                                          & (\underline{79.0}, \underline{52.7}, \textbf{68.6})           & (64.9, 26.2, 43.8)                                        & (\textbf{81.3}, \textbf{55.8}, \underline{56.8})   \\
                                                                                         & MVTec    & (80.9, 48.2,   23.1)                                            & (89.4, 37.8, 24.3)                                            & (\underline{90.0}, \underline{80.2}, \underline{36.4})      & (89.6, 76.9, 21.9)                                            & (66.0, 20.2, 0.3)                                         & (\textbf{90.5}, \textbf{80.9}, \textbf{37.0})   \\
                                                                                         & Visa     & (82.1, 45.2,   3.6)                                             & (90.6, 48.9, \underline{14.1})                                & (\underline{91.2}, \underline{73.9}, 10.9)                  & (90.0, 71.6, 8.7)                                             & (68.8, 21.2, 0.2)                                         & (\textbf{92.0}, \textbf{80.2}, \textbf{23.2})   \\
                                                                                         & MPDD     & (75.0, 22.0,   0.4)                                             & (91.7, 57.5, \underline{15.7})                                & (\underline{91.9}, \underline{80.9}, 9.0)                   & (91.8, 78.2, 5.2)                                             & (81.9, 42.4, 0.9)                                         & (\textbf{93.7}, \textbf{81.7}, \textbf{24.3})   \\
                                                                                         & SDD      & (78.5, 38.3,   \underline{20.4})                                & (84.5, 0.8, 4.2)                                              & (86.2, 53.8, 8.3)                                           & (\underline{86.3}, \underline{55.3}, 8.9)                     & (49.2, 9.5, 0.3)                                          & (\textbf{86.9}, \textbf{65.0}, \textbf{24.4})   \\
                                                                                         & BTAD     & (63.3,   13.1, 5.4)                                             & (84.4, 7.2, 15.9)                                             & (\textbf{87.0}, \textbf{67.6},   \underline{25.4})            & (79.9, 49.7, 11.9)                                            & (52.3, 13.4, 4.3)                                         & (\underline{85.7}, \underline{60.6}, \textbf{27.2})   \\ \cline{2-8} 
                                                                                         & Average  & (72.9, 32.5,   16.3)                                            & (84.1, 24.1, 28.1)                                            & (82.3, \underline{65.5}, 29.4)                              & (\underline{84.4}, 64.9, \underline{34.3})                    & (63.9, 24.6, 17.2)                                        & (\textbf{87.3}, \textbf{71.1}, \textbf{40.4})   \\ \hline
\end{tabular}
}
\end{table}

\begin{table}[t]
\caption{Performance comparison of DIVE with baseline models utilizing MVTec as auxiliary data for pre-training.}
\label{table2}
\scalebox{0.63}{
\begin{tabular}{c|c|cccccc}
\hline
Task                                                                                     & Dataset  & \begin{tabular}[c]{@{}c@{}}AnomalyCLIP\\ (ICLR' 2024)\end{tabular} & \begin{tabular}[c]{@{}c@{}}AdaCLIP\\ (ECCV' 2024)\end{tabular} & \begin{tabular}[c]{@{}c@{}}AF-CLIP\\ (MM' 2025)\end{tabular} & \begin{tabular}[c]{@{}c@{}}AA-CLIP\\ (CVPR' 2025)\end{tabular} & \begin{tabular}[c]{@{}c@{}}TPS\\ (AAAI' 2025)\end{tabular} & DIVE               \\ \hline
\multirow{9}{*}{\begin{tabular}[c]{@{}c@{}}Classification\\ (image-level)\end{tabular}}  & HeadCT   & (\underline{98.3}, 77.9)                                        & (97.4, 78.0)                                                  & (92.9, 69.9)                                                & (87.5, 39.2)                                                  & (98.1, \underline{82.4})                                  & (\textbf{99.1}, \textbf{91.2})   \\
                                                                                         & Br35H    & (95.1, 71.8)                                                    & (95.4, 72.6)                                                  & (\underline{96.0}, \underline{75.4})                        & (82.5, 55.4)                                                  & (95.2, 73.4)                                              & (\textbf{96.1}, \textbf{76.9})   \\
                                                                                         & BrainMRI & (95.5, 77.2)                                                    & (91.2, 71.3)                                                  & (\underline{96.5}, \underline{79.0})                        & (84.9, 52.7)                                                  & (94.1, 77.4)                                              & (\textbf{97.3}, \textbf{84.0})   \\
                                                                                         & DTD      & (91.5, 78.3)                                                    & (\textbf{97.6}, \textbf{92.1})                                & (89.6, 75.4)                                                & (86.6, 77.2)                                                  & (91.3, 81.4)                                              & (\underline{92.8}, \underline{84.0}) \\
                                                                                         & Visa     & (81.7, 53.7)                                                    & (83.4, 57.4)                                                  & (\textbf{85.0}, \textbf{61.8})                              & (73.9, 43.5)                                                  & (83.3, 56.4)                                              & (\underline{84.0}, \underline{57.7}) \\
                                                                                         & MPDD     & (72.2, 31.4)                                                    & (\underline{75.1}, 38.8)                                      & (74.0, \underline{40.8})                                    & (59.7, 23.3)                                                  & (73.1, 31.9)                                              & (\textbf{75.6}, \textbf{41.1})   \\
                                                                                         & SDD      & (83.0, 70.1)                                                    & (81.6, 59.5)                                                  & (73.7, 45.9)                                                & (71.6, 40.7)                                                  & (\underline{83.8}, \textbf{73.6})                         & (\textbf{83.9}, \underline{71.1}) \\
                                                                                         & BTAD     & (91.5, 87.1)                                                    & (90.9, \underline{90.2})                                      & (\underline{91.8}, 89.1)                                    & (82.2, 85.4)                                                  & (90.7, 86.1)                                              & (\textbf{92.0}, \textbf{90.4})   \\ \cline{2-8} 
                                                                                         & Average  & (88.6, 68.4)                                                    & (\underline{89.1}, 70.0)                                      & (87.4, 67.2)                                                & (78.6, 52.2)                                                  & (88.7, \underline{70.3})                                  & (\textbf{90.1}, \textbf{74.6})   \\ \hline
\multirow{9}{*}{\begin{tabular}[c]{@{}c@{}}Segmentation\\ (pixel-level)\end{tabular}}    & ISIC     & (84.9, 69.4, 69.6)                                              & (81.4, 13.9, 69.4)                                            & (\textbf{90.1}, \textbf{78.1}, \textbf{85.7})               & (77.6, 68.4, \underline{85.4})                                & (66.4, 40.0, 59.4)                                        & (\underline{88.5}, \underline{76.4}, 73.9) \\
                                                                                         & ColonDB  & (\underline{82.1}, \underline{69.7}, 53.4)                      & (80.8, 21.2, 36.7)                                            & (80.7, 66.6, 51.2)                                          & (73.4, 63.0, \underline{54.3})                                & (60.1, 25.5, 28.1)                                        & (\textbf{83.2}, \textbf{71.4}, \textbf{54.9}) \\
                                                                                         & TN3K     & (\underline{82.0}, 49.3, 51.2)                                  & (78.4, 5.8, 52.8)                                             & (81.4, \textbf{62.2}, \textbf{62.7})                        & (67.1, 35.5, 52.4)                                            & (65.5, 28.9, 45.1)                                        & (\textbf{84.1}, \underline{54.5}, \underline{55.7}) \\
                                                                                         & DTD      & (98.6, \underline{92.8}, \underline{48.1})                      & (99.0, 65.9, 27.9)                                            & (\underline{99.5}, 92.0, 37.0)                              & (92.9, 81.5, 33.3)                                            & (86.9, 65.5, 11.1)                                        & (\textbf{99.7}, \textbf{94.7}, \textbf{49.9}) \\
                                                                                         & Visa     & (94.9, \underline{81.8}, \underline{24.5})                      & (92.6, 33.9, 16.2)                                            & (\textbf{96.2}, 81.1, 23.9)                                 & (92.7, 76.8, 10.9)                                            & (79.4, 44.7, 0.6)                                         & (\underline{95.0}, \textbf{85.1}, \textbf{25.4}) \\
                                                                                         & MPDD     & (92.3, 77.8, 17.8)                                              & (94.5, 24.3, 17.7)                                            & (\underline{95.3}, 82.4, \textbf{20.4})                     & (94.5, \underline{82.5}, 7.6)                                 & (87.5, 58.6, 0.2)                                         & (\textbf{95.5}, \textbf{84.4}, \underline{18.5}) \\
                                                                                         & SDD      & (\underline{91.3}, \underline{59.9}, \underline{32.6})          & (86.0, 35.2, 26.8)                                            & (\textbf{90.5}, 55.5, 26.0)                                 & (77.0, 44.5, 9.9)                                             & (72.6, 33.8, 6.2)                                         & (88.2, \textbf{60.0}, \textbf{33.5}) \\
                                                                                         & BTAD     & (93.8, \underline{72.9}, 34.0)                                  & (93.6, 47.9, 29.4)                                            & (\textbf{95.0}, 53.8, \underline{34.6})                     & (91.3, 71.7, 20.3)                                            & (66.4, 29.9, 6.1)                                         & (\underline{94.3}, \textbf{73.6}, \textbf{37.6}) \\ \cline{2-8} 
                                                                                         & Average  & (\underline{90.0}, \underline{71.7}, 41.4)                      & (88.3, 31.0, 34.6)                                            & (\textbf{91.1}, 71.5, \underline{42.7})                     & (83.3, 65.5, 34.3)                                            & (73.1, 40.9, 19.6)                                        & (\textbf{91.1}, \textbf{75.0}, \textbf{43.7}) \\ \hline
\end{tabular}
}
\end{table}

\subsubsection{Implementation Details.} The pre-trained CLIP model (ViT-L/14@336px) was adopted as our backbone, with all input images uniformly resized to a resolution of $518 \times 518$. The learnable parameters detailed in Section~\ref{Training and Inference} were updated using the Adam optimizer. Training was conducted with a batch size of 8, applying a learning rate of $1 \times 10^{-3}$ to the prompt learners and coupling function $\mathcal{F}$, and $5 \times 10^{-5}$ to the cross-attention network $\mathcal{CA}$. Within the text encoder, the learnable prompt length was set to $m=12$ and the prompting depth to $P=9$. The length of the text prompts used for projection was configured to $h=4$. For the vision encoder, cross-attention modules were introduced into the final four ViT blocks (i.e., from $Q'=21$ to $Q=24$) to enable the injection of the LLM-generated text embeddings. For all baseline models, we used their official implementations released by the authors. All experiments were implemented in PyTorch 2.0.0 and conducted on a server equipped with a single NVIDIA L40S GPU, an Intel(R) Xeon(R) Silver 4110 CPU, and 160 GB of RAM.

\subsection{Experimental Results and Analysis}
\subsubsection{Overall Performance.}
Comparing Table~\ref{table1} and Table~\ref{table2} reveals a significant performance degradation when utilizing DTD as auxiliary data instead of MVTec. In the image-level classification task, baseline models suffer an average drop of 7.1\%-11.6\% in AUROC and 9.9\%-23.3\% in AP, with a similar downward trend consistently observed in the segmentation task. This discrepancy arises because MVTec provides diverse anomaly classes that naturally share closer patterns with the target domains. Since real-world applications frequently encounter unpredictable domain shifts, the DTD pre-training setting presents a more realistic and challenging scenario. In contrast, DIVE successfully mitigates this degradation under this challenging setting, restricting the performance drop in the classification task to a mere 2.8\% for AUROC and 5.3\% for AP. DIVE yields the best results on nearly all testing datasets and ranks second-best on the rare exceptions. For image-level classification, DIVE achieves average improvements of 5.7\%-16.2\% in AUROC and 8.8\%-28.5\% in AP. Similarly, for pixel-level segmentation, it delivers substantial average gains of 2.9\%-23.4\% in AUROC, 5.6\%-47.0\% in AUPRO, and 6.1\%-24.1\% in our proposed RCPRO metric. Furthermore, Table~\ref{table2} also demonstrates that given sufficient anomaly diversity in the auxiliary data, DIVE maintains highly competitive performance, yielding the best average results across all evaluated datasets.

\begin{figure*}[!t]
\centering
\includegraphics[width=10cm]{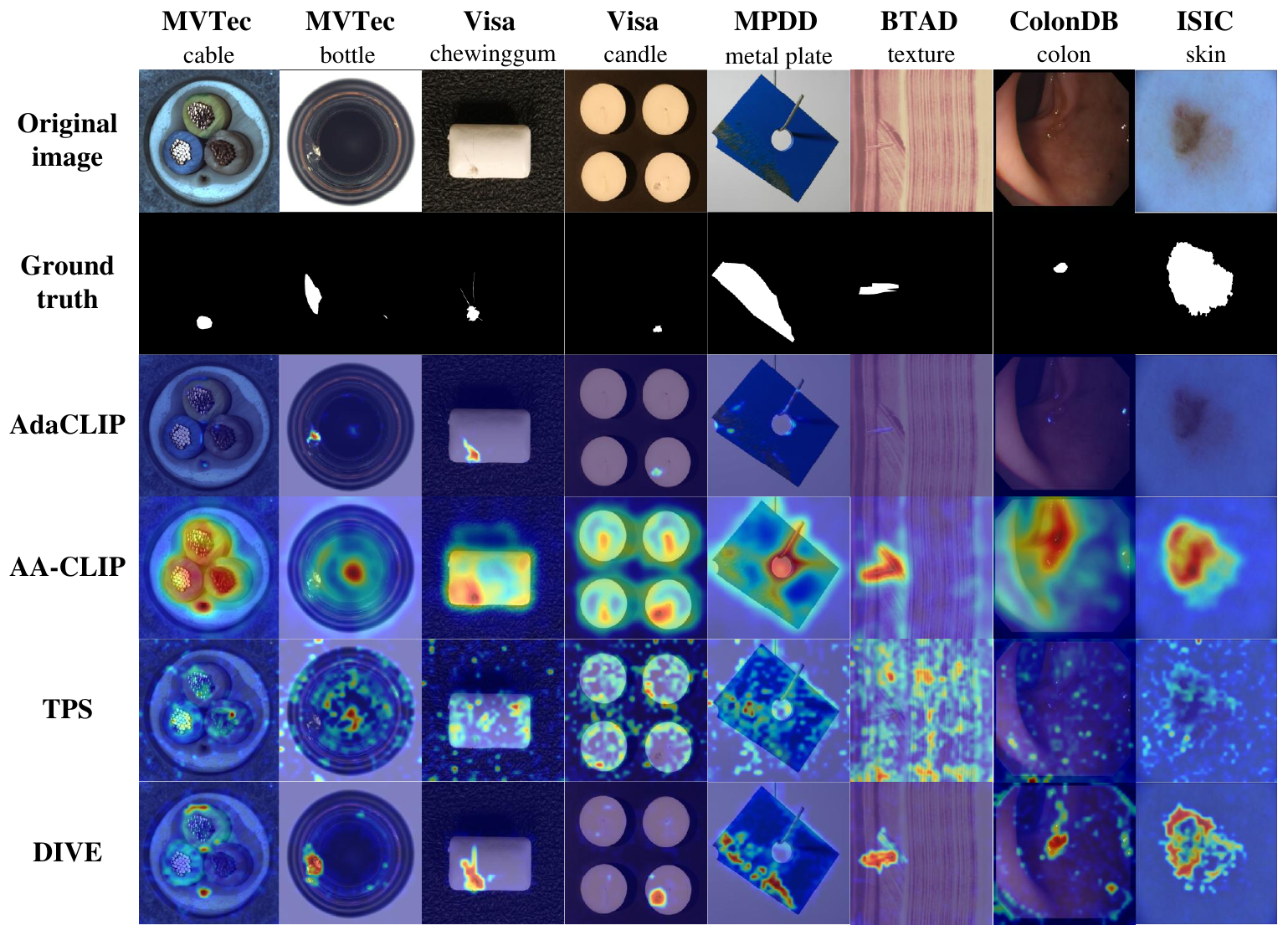}
\caption{Visualization of anomaly maps generated by DIVE and baseline models across diverse datasets, under the setting where DTD is utilized as auxiliary pre-training data.}
\label{visualization}
\end{figure*}

To delve into several intriguing phenomena observed in Table~\ref{table1}, Fig.~\ref{visualization} visualizes the anomaly maps generated by DIVE and three relevant baselines for further analysis. \textit{Observation 1.} Although TPS ranks second or third in image-level classification across most datasets, it underperforms in segmentation. As illustrated by its anomaly maps, although TPS possesses the capacity to recognize anomalous images globally, it generates numerous small-scale false positives scattered across normal regions. \textit{Observation 2.} AdaCLIP outperforms AA-CLIP in our proposed RCPRO metric on several industrial datasets (e.g., MVTec, Visa, and MPDD), despite falling significantly behind in AUPRO. The visualization results show that AA-CLIP tends to predict excessively large connected components, leading to severe over-segmentation and prominent false positives in normal regions. In contrast, AdaCLIP can localize the core anomalies, albeit with potentially incomplete boundaries. As the AUPRO metric primarily emphasizes the recall of ground-truth regions without imposing penalties for extraneous predictions, AA-CLIP receives an inflated AUPRO score. This phenomenon compellingly demonstrates the necessity and rationality of our RCPRO metric. \textit{Observation 3.} AA-CLIP yields higher RCPRO scores than DIVE on several medical imaging datasets, most notably ISIC. This exception stems from the specific characteristics of the ISIC dataset, where anomalous regions occupy a dominant portion of the entire image. Consequently, AA-CLIP’s tendency to predict excessively expansive connected regions, which causes over-segmentation elsewhere, inadvertently becomes a advantage here. Nevertheless, DIVE still precisely segments the core pathological areas of these anomalies.

\begin{table}[t]
\caption{Performance of DIVE and its ablated variants on the MVTec and Visa datasets. The evaluation metrics for the classification task are (AUROC, AP), and for the segmentation task are (AUROC, AUPRO, RCPRO).}
\centering
\label{ablation table}
\scalebox{0.78}{
\begin{tabular}{c|cc|cc}
\hline
\multirow{2}{*}{Models} & \multicolumn{2}{c|}{MVTec}          & \multicolumn{2}{c}{Visa}            \\ \cline{2-5} 
                        & Classification & Segmentation       & Classification & Segmentation       \\ \hline
DIVE$_{\text{-dis}}$         & (88.7$_{\downarrow 0.6}$, 72.1$_{\downarrow 1.8}$)   & (89.9$_{\downarrow 0.6}$, 80.5$_{\downarrow 0.4}$, 36.8$_{\downarrow 0.2}$) & (76.2$_{\downarrow 1.6}$, 48.4$_{\downarrow 1.6}$)   & (91.8$_{\downarrow 0.2}$, 79.5$_{\downarrow 0.7}$, 22.9$_{\downarrow 0.3}$) \\ \hline
DIVE$_{\text{-text(s)}}$      & (88.8$_{\downarrow 0.5}$, 73.2$_{\downarrow 0.7}$)   & (90.2$_{\downarrow 0.3}$, 80.5$_{\downarrow 0.4}$, 36.2$_{\downarrow 0.8}$) & (76.9$_{\downarrow 0.9}$, 49.1$_{\downarrow 0.9}$)   & (91.6$_{\downarrow 0.4}$, 79.4$_{\downarrow 0.8}$, 23.1$_{\downarrow 0.1}$) \\
DIVE$_{\text{-text(d)}}$      & (87.3$_{\downarrow 2.0}$, 70.8$_{\downarrow 3.1}$)   & (89.7$_{\downarrow 0.8}$, 77.3$_{\downarrow 3.6}$, 35.6$_{\downarrow 1.4}$) & (75.3$_{\downarrow 2.5}$, 46.0$_{\downarrow 4.0}$)   & (91.1$_{\downarrow 0.9}$, 77.9$_{\downarrow 2.3}$, 21.6$_{\downarrow 1.6}$) \\
DIVE$_{\text{-text}}$         & (86.9$_{\downarrow 2.4}$, 69.5$_{\downarrow 4.4}$)   & (88.3$_{\downarrow 2.2}$, 76.6$_{\downarrow 4.3}$, 35.5$_{\downarrow 1.5}$) & (74.4$_{\downarrow 3.4}$, 43.9$_{\downarrow 6.1}$)   & (90.4$_{\downarrow 1.6}$, 74.4$_{\downarrow 5.8}$, 20.5$_{\downarrow 2.7}$) \\ \hline
DIVE                    & (89.3, 73.9)   & (90.5, 80.9, 37.0) & (77.8, 50.0)   & (92.0, 80.2, 23.2) \\ \hline
\end{tabular}
}
\end{table}

\subsubsection{Ablation Study.} To evaluate the contribution of each proposed component, we perform ablation studies using several DIVE variants: 1) DIVE$_{\text{-dis}}$ ablates the disentanglement mechanism; 2) DIVE$_{\text{-text}}$ entirely discards the text embedding injection.  DIVE$_{\text{-text(s)}}$ and DIVE$_{\text{-text(d)}}$ are further introduced to ablate the shallow-level and deep-level injections, respectively. Table~\ref{ablation table} presents the results of the aforementioned ablated models under the setting of using DTD as auxiliary data. Comparing DIVE$_{\text{-dis}}$ with the full DIVE model reveals that the performance degradation predominantly occurs in the two classification metrics. This is attributed to the fact that the disentanglement operates on the global visual embeddings, which are aligned with state textual embeddings to derive the image-level anomaly scores. Furthermore, we observe that the performance degradation of DIVE$_{\text{-text}}$ is more pronounced than that of DIVE$_{\text{-dis}}$. In scenarios where the auxiliary data lacks sufficient anomaly diversity, the text embedding injection strategy is critical to capture generic anomaly concepts, thereby safeguarding the model against a decline in generalization.

\begin{figure*}[!t]
\centering
\includegraphics[width=12.2cm]{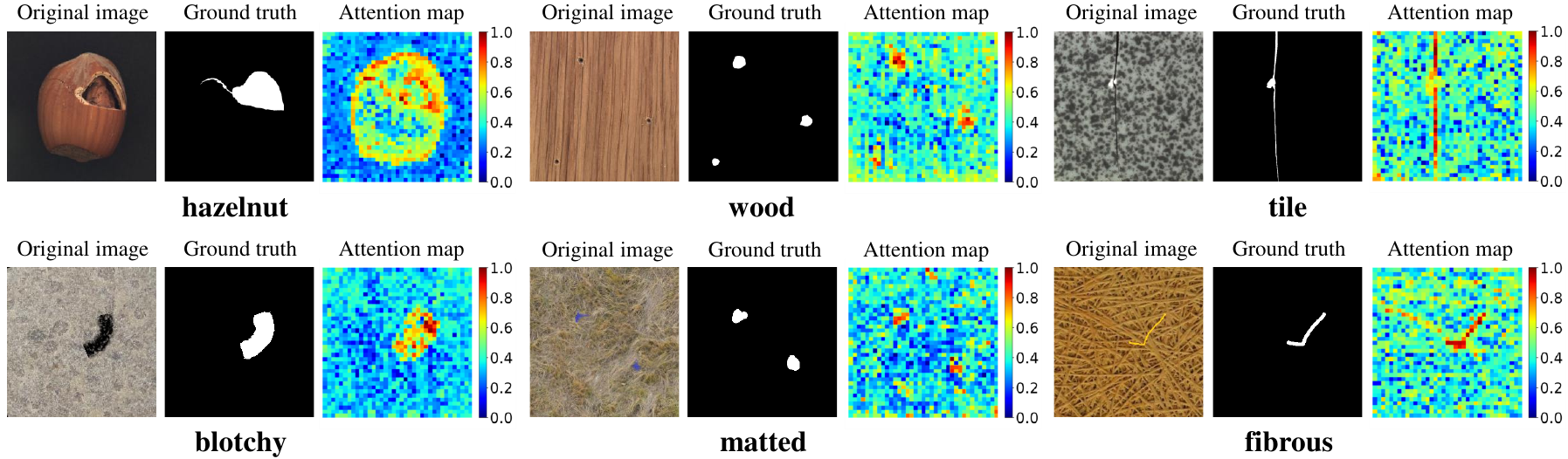}
\caption{Visualization of attention maps between image patches of representative categories from the auxiliary data (first row: MVTec, second row: DTD) and the embeddings of LLM-generated descriptions. The attention score for each patch is calculated by summing the normalized similarity probabilities of all anomaly descriptions.}
\label{attention}
\end{figure*}

A further analysis of the DIVE$_{\text{-text}}$ variants shows that DIVE$_{\text{-text(d)}}$ exhibits a more pronounced performance degradation than DIVE$_{\text{-text(s)}}$, underscoring the importance of incorporating LLM-generated descriptions of normality and abnormality into the deep-level patch tokens. Within the DIVE framework, the cross-attention networks are tasked with integrating these text embeddings, and the resulting patch-wise attention maps are visualized in Fig.~\ref{attention}. It can be observed that the cross-attention module in DIVE excels at focusing on anomalous regions, though it also exhibits a slight tendency to attend to object boundaries. For instance, regarding the hazelnut category in Fig.~\ref{attention}, high attention scores are assigned not only to the damaged area but also to the contour of the hazelnut against the background. Consequently, as shown in Fig.~\ref{visualization}, DIVE may occasionally introduce minor false positive predictions along object edges (e.g., the anomaly maps for the cable in MVTec and the colon in ColonDB).

\noindent
\subsubsection{Hyperparameter Sensitivity.} 
\begin{figure*}[!t]
\centering
\includegraphics[width=12cm]{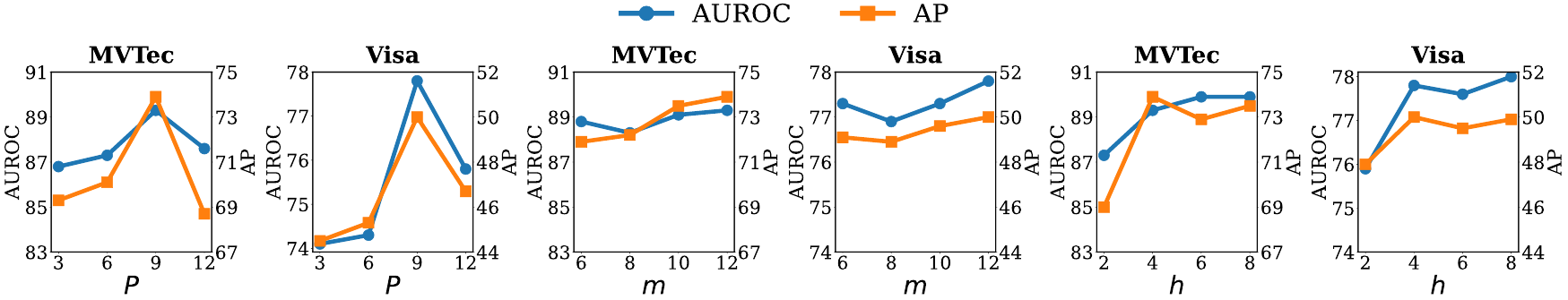}
\caption{AUROC ((left Y-axis) and AP (right Y-axis) results under varying values of the hyperparameters $P$, $m$, and $h$.}
\label{para}
\end{figure*}

In this section, we conduct a sensitivity analysis of DIVE on the MVTec and Visa datasets concerning three hyperparameters: the prompting depth $P$, the learnable prompt length $m$, and the length of the text prompts used for projection $h$. First, to evaluate the model's sensitivity to $P$, we keep the other two parameters fixed and vary the value of $P$ within the set $\{3, 6, 9, 12\}$. As illustrated in Fig.~\ref{para}, DIVE is relatively sensitive to the parameter $P$. With the increase of $P$, the model's performance exhibits an initial upward trend followed by a subsequent decline. This phenomenon can be attributed to the fact that deeper prompt layers facilitate a more refined construction of the textual embedding space regarding anomalous states and object semantics; however, when the prompting layers become excessively deep, it may degrade the original representation capacity of the pre-trained VLMs.

Next, we evaluate the sensitivity of DIVE to the hyperparameters $m$ and $h$, with their variation ranges set to $\{6, 8, 10, 12\}$ and $\{2, 4, 6, 8\}$, respectively. The hyperparameter $m$ denotes the length of the learnable context initialized at the first layer of the text encoder. As observed in Fig.~\ref{para}, DIVE demonstrates relatively low sensitivity to $m$. This can be interpreted as indicating that a small number of learnable context tokens is already sufficient to adequately capture the essential information required to distinguish between normal and anomalous states. Regarding the hyperparameter $h$, we select a relatively small value (e.g., $4$). This choice is motivated by the observation that the DIVE's performance tends to plateau beyond this value. Furthermore, maintaining a smaller value prevents the visual encoding process from incurring excessive overhead or introducing redundant noise.

\section{Conclusion}
In this paper, we investigate a more realistic and challenging scenario for ZSAD characterized by limited auxiliary anomaly priors. To address this, we propose DIVE, a novel model that integrates and fine-tunes learnable networks and prompts to construct robust visual embeddings. By leveraging text embedding injection and a disentanglement mechanism, DIVE is enable to capture anomaly patterns shared between the auxiliary training domain and unknown target domains. Extensive experiments demonstrate that DIVE exhibiting superior generalization on testing data that heavily deviates from the auxiliary training distribution. Future work will focus on mitigating the small false-positive noise patches introduced by DIVE along object boundaries, to further enhance its fine-grained anomaly localization capability.

%
%
\bibliographystyle{splncs04}
\bibliography{main}

@String(CVPR  = {IEEE Conf. Comput. Vis. Pattern Recog.})

@String(ICCV  = {Int. Conf. Comput. Vis.})

@String(ECCV  = {Eur. Conf. Comput. Vis.})

@String(ICML  = {Int. Conf. Mach. Learn.})

@String(ICLR  = {Int. Conf. Learn. Represent.})

@String(AAAI  = {AAAI})

@String(IJCAI = {IJCAI})

@String(CVPR  = {CVPR})

@String(ICCV  = {ICCV})

@String(ECCV  = {ECCV})

@String(ICML  = {ICML})

@String(ICLR  = {ICLR})

@inproceedings{cao2024adaclip,
  title={Adaclip: Adapting clip with hybrid learnable prompts for zero-shot anomaly detection},
  author={Cao, Yunkang and Zhang, Jiangning and Frittoli, Luca and Cheng, Yuqi and Shen, Weiming and Boracchi, Giacomo},
  booktitle={ECCV},
  pages={55--72},
  year={2024}
}

@inproceedings{zhou2023anomalyclip,
  title={Anomalyclip: Object-agnostic prompt learning for zero-shot anomaly detection},
  author={Zhou, Qihang and Pang, Guansong and Tian, Yu and He, Shibo and Chen, Jiming},
  booktitle={ICLR},
  year={2024}
}

@inproceedings{ma2025aligning,
  title={Aligning and prompting anything for zero-shot generalized anomaly detection},
  author={Ma, Jitao and Xie, Weiying and Ye, Hangyu and Li, Daixun and Fang, Leyuan},
  booktitle={AAAI},
  volume={39},
  number={6},
  pages={5964--5972},
  year={2025}
}

@inproceedings{ma2025aa,
  title={Aa-clip: Enhancing zero-shot anomaly detection via anomaly-aware clip},
  author={Ma, Wenxin and Zhang, Xu and Yao, Qingsong and Tang, Fenghe and Wu, Chenxu and Li, Yingtai and Yan, Rui and Jiang, Zihang and Zhou, S Kevin},
  booktitle={CVPR},
  pages={4744--4754},
  year={2025}
}

@inproceedings{khattak2023maple,
  title={Maple: Multi-modal prompt learning},
  author={Khattak, Muhammad Uzair and Rasheed, Hanoona and Maaz, Muhammad and Khan, Salman and Khan, Fahad Shahbaz},
  booktitle={CVPR},
  pages={19113--19122},
  year={2023}
}

@inproceedings{lin2017focal,
  title={Focal loss for dense object detection},
  author={Lin, Tsung-Yi and Goyal, Priya and Girshick, Ross and He, Kaiming and Doll{\'a}r, Piotr},
  booktitle={ICCV},
  pages={2980--2988},
  year={2017}
}

@inproceedings{li2020dice,
  title={Dice loss for data-imbalanced NLP tasks},
  author={Li, Xiaoya and Sun, Xiaofei and Meng, Yuxian and Liang, Junjun and Wu, Fei and Li, Jiwei},
  booktitle={ACL},
  pages={465--476},
  year={2020}
}

@inproceedings{zhu2025fine,
  title={Fine-grained abnormality prompt learning for zero-shot anomaly detection},
  author={Zhu, Jiawen and Ong, Yew-Soon and Shen, Chunhua and Pang, Guansong},
  booktitle={ICCV},
  pages={22241--22251},
  year={2025}
}

@inproceedings{radford2021learning,
  title={Learning transferable visual models from natural language supervision},
  author={Radford, Alec and Kim, Jong Wook and Hallacy, Chris and Ramesh, Aditya and Goh, Gabriel and Agarwal, Sandhini and Sastry, Girish and Askell, Amanda and Mishkin, Pamela and Clark, Jack and others},
  booktitle={ICML},
  pages={8748--8763},
  year={2021}
}

@inproceedings{jeong2023winclip,
  title={Winclip: Zero-/few-shot anomaly classification and segmentation},
  author={Jeong, Jongheon and Zou, Yang and Kim, Taewan and Zhang, Dongqing and Ravichandran, Avinash and Dabeer, Onkar},
  booktitle={CVPR},
  pages={19606--19616},
  year={2023}
}

@article{chen2023zero,
  title={A Zero-/Few-Shot Anomaly Classification and Segmentation Method for CVPR 2023 VAND Workshop Challenge Tracks 1\&2: 1st Place on Zero-shot AD and 4th Place on Few-shot AD},
  author={Chen, Xuhai and Han, Yue and Zhang, Jiangning},
  journal={arXiv preprint arXiv:2305.17382},
  year={2023}
}

@article{anovl,
  title={AnoVL: Adapting Vision-Language Models for Unified Zero-shot Anomaly Localization},
  author={Deng, Hanqiu and Zhang, Zhaoxiang and Bao, Jinan and Li, Xingyu},
  journal={arXiv preprint arXiv:2308.15939},
  year={2023}
}

@inproceedings{fang2025af,
  title={Af-clip: Zero-shot anomaly detection via anomaly-focused clip adaptation},
  author={Fang, Qingqing and Lv, Wenxi and Su, Qinliang},
  booktitle={ACM MM},
  pages={4846--4855},
  year={2025}
}

@article{cao2024survey,
  title={A survey on visual anomaly detection: Challenge, approach, and prospect},
  author={Cao, Yunkang and Xu, Xiaohao and Zhang, Jiangning and Cheng, Yuqi and Huang, Xiaonan and Pang, Guansong and Shen, Weiming},
  journal={arXiv preprint arXiv:2401.16402},
  year={2024}
}

@article{li2025survey,
  title={A survey of deep learning for industrial visual anomaly detection},
  author={Li, Zhuo and Yan, Yuhao and Wang, Xiangheng and Ge, Yifei and Meng, Lin},
  journal={Artif. Intell. Rev.},
  volume={58},
  number={9},
  pages={279},
  year={2025}
}

@inproceedings{zhu2024toward,
  title={Toward generalist anomaly detection via in-context residual learning with few-shot sample prompts},
  author={Zhu, Jiawen and Pang, Guansong},
  booktitle={CVPR},
  pages={17826--17836},
  year={2024}
}

@inproceedings{ho2024long,
  title={Long-tailed anomaly detection with learnable class names},
  author={Ho, Chih-Hui and Peng, Kuan-Chuan and Vasconcelos, Nuno},
  booktitle={CVPR},
  pages={12435--12446},
  year={2024}
}

@inproceedings{cai2024rethinking,
  title={Rethinking autoencoders for medical anomaly detection from a theoretical perspective},
  author={Cai, Yu and Chen, Hao and Cheng, Kwang-Ting},
  booktitle={MICCAI},
  pages={544--554},
  year={2024}
}

@inproceedings{zhang2024mediclip,
  title={Mediclip: Adapting clip for few-shot medical image anomaly detection},
  author={Zhang, Ximiao and Xu, Min and Qiu, Dehui and Yan, Ruixin and Lang, Ning and Zhou, Xiuzhuang},
  booktitle={MICCAI},
  pages={458--468},
  year={2024}
}

@inproceedings{lu2024targeted,
  title={Targeted Detection of Anomalous Merchants on Integrated Payment Platforms via Multifaceted Transaction Representation Learning},
  author={Lu, Guanyu and Lin, Xiang and Pavlovski, Martin and Zhang, Xinyu and Zhou, Fang},
  booktitle={IEEE BigData},
  pages={2170--2178},
  year={2024}
}

@inproceedings{bergmann2019mvtec,
  title={MVTec AD--A comprehensive real-world dataset for unsupervised anomaly detection},
  author={Bergmann, Paul and Fauser, Michael and Sattlegger, David and Steger, Carsten},
  booktitle={CVPR},
  pages={9592--9600},
  year={2019}
}

@inproceedings{bergmann2020uninformed,
  title={Uninformed students: Student-teacher anomaly detection with discriminative latent embeddings},
  author={Bergmann, Paul and Fauser, Michael and Sattlegger, David and Steger, Carsten},
  booktitle={CVPR},
  pages={4183--4192},
  year={2020}
}

@inproceedings{gong2025fe,
  title={FE-CLIP: Frequency Enhanced CLIP Model for Zero-Shot Anomaly Detection and Segmentation},
  author={Gong, Tao and Chu, Qi and Liu, Bin and Zhou, Wei and Yu, Nenghai},
  booktitle={ICCV},
  pages={21220--21230},
  year={2025}
}

@inproceedings{he2025rareclip,
  title={Rareclip: Rarity-aware online zero-shot industrial anomaly detection},
  author={He, Jianfang and Cao, Min and Peng, Silong and Xie, Qiong},
  booktitle={ICCV},
  pages={24478--24487},
  year={2025}
}

@inproceedings{aota2023zero,
  title={Zero-shot versus many-shot: Unsupervised texture anomaly detection},
  author={Aota, Toshimichi and Tong, Lloyd Teh Tzer and Okatani, Takayuki},
  booktitle={WACV},
  pages={5564--5572},
  year={2023}
}

@inproceedings{zou2022spot,
  title={Spot-the-difference self-supervised pre-training for anomaly detection and segmentation},
  author={Zou, Yang and Jeong, Jongheon and Pemula, Latha and Zhang, Dongqing and Dabeer, Onkar},
  booktitle={ECCV},
  pages={392--408},
  year={2022}
}

@inproceedings{jezek2021deep,
  title={Deep learning-based defect detection of metal parts: evaluating current methods in complex conditions},
  author={Jezek, Stepan and Jonak, Martin and Burget, Radim and Dvorak, Pavel and Skotak, Milos},
  booktitle={ICUMT},
  pages={66--71},
  year={2021}
}

@article{tabernik2020segmentation,
  title={Segmentation-based deep-learning approach for surface-defect detection},
  author={Tabernik, Domen and {\v{S}}ela, Samo and Skvar{\v{c}}, Jure and Sko{\v{c}}aj, Danijel},
  journal={J. Intell. Manuf.},
  volume={31},
  number={3},
  pages={759--776},
  year={2020}
}

@inproceedings{Mishra2021VTADLAV,
  title={VT-ADL: A Vision Transformer Network for Image Anomaly Detection and Localization},
  author={Pankaj Mishra and Riccardo Verk and Daniele Fornasier and Claudio Piciarelli and Gian Luca Foresti},
  booktitle={ISIE},
  year={2021},
  pages={01-06}
}

@inproceedings{salehi2021multiresolution,
  title={Multiresolution knowledge distillation for anomaly detection},
  author={Salehi, Mohammadreza and Sadjadi, Niousha and Baselizadeh, Soroosh and Rohban, Mohammad H and Rabiee, Hamid R},
  booktitle={CVPR},
  pages={14902--14912},
  year={2021}
}

@misc{hamada2020br35h,
  author = {Ahmed Hamada},
  title  = {Br35H: Brain Tumor Detection 2020},
  year   = {2020},
  howpublished = {\url{https://www.kaggle.com/datasets/ahmedhamada0/brain-tumor-detection}},
  note   = {Kaggle dataset}
}

@article{gutman2016skin,
  title={Skin lesion analysis toward melanoma detection: A challenge at the international symposium on biomedical imaging (ISBI) 2016, hosted by the international skin imaging collaboration (ISIC)},
  author={Gutman, David and Codella, Noel CF and Celebi, Emre and Helba, Brian and Marchetti, Michael and Mishra, Nabin and Halpern, Allan},
  journal={arXiv preprint arXiv:1605.01397},
  year={2016}
}

@article{tajbakhsh2015automated,
  title={Automated polyp detection in colonoscopy videos using shape and context information},
  author={Tajbakhsh, Nima and Gurudu, Suryakanth R and Liang, Jianming},
  journal={IEEE Trans. Med. Imaging},
  volume={35},
  number={2},
  pages={630--644},
  year={2015}
}

@inproceedings{gong2021multi,
  title={Multi-task learning for thyroid nodule segmentation with thyroid region prior},
  author={Gong, Haifan and Chen, Guanqi and Wang, Ranran and Xie, Xiang and Mao, Mingzhi and Yu, Yizhou and Chen, Fei and Li, Guanbin},
  booktitle={ISBI},
  pages={257--261},
  year={2021}
}

@inproceedings{xu2023fascinating,
  title={Fascinating supervisory signals and where to find them: Deep anomaly detection with scale learning},
  author={Xu, Hongzuo and Wang, Yijie and Wei, Juhui and Jian, Songlei and Li, Yizhou and Liu, Ning},
  booktitle={ICML},
  pages={38655--38673},
  year={2023}
}

@article{xu2023deep,
  title={Deep isolation forest for anomaly detection},
  author={Xu, Hongzuo and Pang, Guansong and Wang, Yijie and Wang, Yongjun},
  journal={IEEE Trans. Knowl. Data Eng.},
  volume={35},
  number={12},
  pages={12591--12604},
  year={2023}
}

@article{zhou2024msflow,
  title={Msflow: Multiscale flow-based framework for unsupervised anomaly detection},
  author={Zhou, Yixuan and Xu, Xing and Song, Jingkuan and Shen, Fumin and Shen, Heng Tao},
  journal={IEEE Trans. Neural Netw. Learn. Syst.},
  year={2024},
}

@inproceedings{guo2025dinomaly,
  title={Dinomaly: The less is more philosophy in multi-class unsupervised anomaly detection},
  author={Guo, Jia and Lu, Shuai and Zhang, Weihang and Chen, Fang and Li, Huiqi and Liao, Hongen},
  booktitle={CVPR},
  pages={20405--20415},
  year={2025}
}

@inproceedings{lu2024robust,
  title={A robust prioritized anomaly detection when not all anomalies are of primary interest},
  author={Lu, Guanyu and Zhou, Fang and Pavlovski, Martin and Zhou, Chenyi and Jin, Cheqing},
  booktitle={ICDE},
  pages={775--788},
  year={2024}
}

@inproceedings{wei2024gad,
  title={GAD: A Generalized Framework for Anomaly Detection at Different Risk Levels},
  author={Wei, Rulan and He, Zewei and Pavlovski, Martin and Zhou, Fang},
  booktitle={CIKM},
  pages={2513--2522},
  year={2024}
}

@inproceedings{shou2025read,
  title={READ: Robust and Efficient Anomaly Detection under Data Contamination and Limited Supervision},
  author={Shou, Hongzhe and Lu, Guanyu and Pavlovski, Martin and Zhou, Fang},
  booktitle={SIGKDD},
  pages={2586--2596},
  year={2025}
}

@inproceedings{jiang2023anomaly,
  title={Anomaly detection with score distribution discrimination},
  author={Jiang, Minqi and Han, Songqiao and Huang, Hailiang},
  booktitle={SIGKDD},
  pages={984--996},
  year={2023}
}

@inproceedings{zhang2024gpt,
  title={Gpt-4v-ad: Exploring grounding potential of vqa-oriented gpt-4v for zero-shot anomaly detection},
  author={Zhang, Jiangning and He, Haoyang and Chen, Xuhai and Xue, Zhucun and Wang, Yabiao and Wang, Chengjie and Xie, Lei and Liu, Yong},
  booktitle={IJCAI},
  pages={3--16},
  year={2024}
}

@inproceedings{gu2024anomalygpt,
  title={Anomalygpt: Detecting industrial anomalies using large vision-language models},
  author={Gu, Zhaopeng and Zhu, Bingke and Zhu, Guibo and Chen, Yingying and Tang, Ming and Wang, Jinqiao},
  booktitle={AAAI},
  volume={38},
  number={3},
  pages={1932--1940},
  year={2024}
}

@inproceedings{li2024promptad,
  title={Promptad: Learning prompts with only normal samples for few-shot anomaly detection},
  author={Li, Xiaofan and Zhang, Zhizhong and Tan, Xin and Chen, Chengwei and Qu, Yanyun and Xie, Yuan and Ma, Lizhuang},
  booktitle={CVPR},
  pages={16838--16848},
  year={2024}
}

@article{zhou2022learning,
  title={Learning to prompt for vision-language models},
  author={Zhou, Kaiyang and Yang, Jingkang and Loy, Chen Change and Liu, Ziwei},
  journal={Int. J. Comput. Vis.},
  volume={130},
  number={9},
  pages={2337--2348},
  year={2022}
}

@inproceedings{lu2022prompt,
  title={Prompt distribution learning},
  author={Lu, Yuning and Liu, Jianzhuang and Zhang, Yonggang and Liu, Yajing and Tian, Xinmei},
  booktitle={CVPR},
  pages={5206--5215},
  year={2022}
}

@inproceedings{yao2023visual,
  title={Visual-language prompt tuning with knowledge-guided context optimization},
  author={Yao, Hantao and Zhang, Rui and Xu, Changsheng},
  booktitle={CVPR},
  pages={6757--6767},
  year={2023}
}

@inproceedings{chen2022plot,
  title={Plot: Prompt learning with optimal transport for vision-language models},
  author={Chen, Guangyi and Yao, Weiran and Song, Xiangchen and Li, Xinyue and Rao, Yongming and Zhang, Kun},
  booktitle={ICLR},
  year={2023}
}

@inproceedings{zhou2022conditional,
  title={Conditional prompt learning for vision-language models},
  author={Zhou, Kaiyang and Yang, Jingkang and Loy, Chen Change and Liu, Ziwei},
  booktitle={CVPR},
  pages={16816--16825},
  year={2022}
}

\clearpage

\appendix
\section{Appendix}

\subsection{Prompt Template for Generating Descriptions of Normality and Abnormality}
\label{Appendix-A.1}
As detailed in the main manuscript, DIVE utilizes a deep-level text embedding injection strategy to abstract generic anomaly concepts. To acquire rich and diverse textual descriptions of normality and abnormality, we leverage the advanced reasoning capabilities of Large Language Models (LLMs). Specifically, the GPT-4o model is applied to generate these descriptive textual priors. The exact prompt template used to query the model is provided below:
\vspace{0.2cm}

\begin{tcolorbox}[colback=gray!5!white,colframe=gray!75!black,title=Prompt Template for Generating Descriptions]
\textbf{System Prompt:}\\
You are a cross-domain expert in anomaly detection (e.g., industrial manufacturing, medical imaging). Your task is to provide a comprehensive and diverse set of generic visual descriptions for both anomalous (defective) and normal (flawless) states. These descriptions should be abstract enough to apply across various domains.

\textbf{Instruction:}\\
List at least 100 diverse descriptions for the anomalous state (e.g., crack, scratch, stain) and at least 35 descriptions for the normal state (e.g., flawless surface, uniform texture). 

\textbf{Output format:}\\
Please provide the results exactly as a Python dictionary with two keys: ``Abnormal Lexicon'' and ``Normal Lexicon''. Each value must be a list of strings. Every single description in the lists must strictly follow this exact syntactic template: ``a photo of an object with [specific state/condition]''.
\end{tcolorbox}

\vspace{0.2cm}
\noindent
Based on the aforementioned prompt, the GPT-4o model generated a comprehensive dictionary containing 100 abnormal and 35 normal descriptions. These descriptions cover a wide range of structural, textural, and color-related states characterizing both normality and abnormality. A truncated example of the standard output format is presented below:

\begin{lstlisting}[language=Python, basicstyle=\ttfamily\small,  breaklines=true, showstringspaces=false]
{
    "Abnormal Lexicon": [
        "a photo of an object with a crack",
        "a photo of an object with a stain",
        "a photo of an object with a color shift",
        "a photo of an object with a texture mismatch",
        ...
    ],
    "Normal Lexicon": [
        "a photo of an object with a flawless surface",
        "a photo of an object with uniform texture",
        "a photo of an object with a clean appearance",
        ...
    ]
}
\end{lstlisting}

\noindent
Due to space constraints, we only list a few representative examples above. The complete set of generated textual descriptions utilized in our experiments is available in our code repository at https://github.com/ZhouF-ECNU/DIVE.

\subsection{Extended Details of Datasets and Baselines}
\label{Appendix-A.2}

\subsubsection{Dataset Statistics.}
Table \ref{tab:dataset_stats} summarizes the detailed statistics of the twelve datasets used in our experiments. Notably, for the MVTec and DTD datasets, a small portion of the training data is partitioned to construct a validation set, which is utilized for hyperparameter tuning for DIVE and the baseline models. To faithfully simulate real-world industrial and clinical scenarios where defects or diseases are inherently rare events, we resampled the datasets to introduce a class imbalance.  Excluding the three medical datasets (ISIC, ColonDB, and TN3K) used for segmentation evaluation, the proportion of anomalous samples in the testing data is constrained to a small minority, ranging from approximately 9.1\% (e.g., Br35H and MPDD) to 18.1\% (e.g., SDD) across the evaluated datasets.

\begin{table*}[htbp]
  \centering
  \caption{Detailed statistics of the twelve datasets evaluated in our experiments. Cls. and Seg. denote whether the dataset provides corresponding annotations and is utilized for anomaly classification (image-level) and anomaly segmentation (pixel-level) tasks, respectively. The symbol `-' indicates that the specific data split or task is not applicable to the dataset in our zero-shot setting.}
  \label{tab:dataset_stats}
  \scalebox{0.9}{
  \begin{tabular}{ccccccccc}
    \toprule
    \multirow{2}{*}{Domain} & \multirow{2}{*}{Dataset} & \multicolumn{2}{c}{Training} & \multicolumn{2}{c}{Testing} & & \multicolumn{2}{c}{Task} \\
    \cmidrule{3-4} \cmidrule{5-6} \cmidrule{8-9}
    & & Normal & Anomaly & Normal & Anomaly & & Cls. & Seg. \\
    \midrule
    \multirow{6}{*}{Industrial} 
    & MVTec    & 4,096      & 423      & 3,269              & 334                 & & \checkmark & \checkmark \\
    & DTD      & 1,007      & 240      & 847              & 120                 & & \checkmark & \checkmark \\
    & Visa     & -     & -     & 9,621              & 1,200                 & & \checkmark & \checkmark \\
    & MPDD     & -     & -     & 1,064              & 107                 & & \checkmark & \checkmark \\
    & SDD      & -     & -     & 181              & 40                 & & \checkmark & \checkmark \\
    & BTAD     & -     & -     & 451              & 90                 & & \checkmark & \checkmark \\
    \midrule
    \multirow{6}{*}{Medical}    
    & HeadCT   & -     & -     & 100              & 10                 & & \checkmark & $\times$   \\
    & Br35H    & -     & -     & 1,500              & 150                 & & \checkmark & $\times$   \\
    & BrainMRI & -     & -     & 98              & 10                 & & \checkmark & $\times$   \\
    & ISIC     & -     & -     & -             & 379                & & $\times$   & \checkmark \\
    & ColonDB  & -     & -     & -             & 380                 & & $\times$   & \checkmark \\
    & TN3K     & -     & -     & -             & 614                 & & $\times$   & \checkmark \\
    \bottomrule
  \end{tabular}
  }
\end{table*}

\subsubsection{Overview of Baseline Models.}
To comprehensively evaluate the performance of our proposed DIVE, we conduct extensive comparisons against five recent state-of-the-art (SOTA) baseline models in the field of zero-shot anomaly detection. These selected competitors include \textbf{AnomalyCLIP} (ICLR' 2024), \textbf{AdaCLIP} (ECCV' 2024), \textbf{AF-CLIP} (MM' 2025), \textbf{AA-CLIP} (CVPR' 2025), and \textbf{TPS} (AAAI' 2025). Brief descriptions of these five baseline models are provided as follows:

\begin{itemize}
    \item \textbf{AnomalyCLIP} proposes an object-agnostic prompt learning framework designed to capture normal and abnormal semantics independent of specific foreground object categories.
    \item \textbf{AdaCLIP} introduces a hybrid prompt learning approach that jointly adapts visual and textual spaces by generating image-aware prompts and imposing token-level attention constraints to achieve precise cross-modal alignment.
    \item \textbf{AF-CLIP} leverages a two-branch prompt learning architecture incorporating a text-driven local anomaly focusing mechanism to simultaneously capture global semantics and selectively enhance fine-grained representations of actual defective regions.
    \item \textbf{AA-CLIP} combines dual-perspective prompting to decouple object semantics from defect appearances, along with a token-adaptive masking mechanism that selectively focuses on critical anomalous cues to sharpen anomaly perception.
    \item \textbf{TPS} addresses the granularity mismatch between anomaly classification and segmentation by generating fine-grained text prompts for precise pixel-level alignment, and leverages a joint learning mechanism to model the complementary dependencies between the two tasks.
\end{itemize}

\subsection{Mathematical Formulation of the RCPRO Metric}
\label{Appendix-A.3}
In this section, we provide the step-by-step mathematical derivation for our newly proposed evaluation metric, RCPRO (Region-Calibrated PRO). To avoid any ambiguity, please note that the mathematical notations defined herein are self-contained and limited to this section, bearing no relation to the symbols defined in the main manuscript.

The final RCPRO metric is formulated by calculating the Area Under the Curve (AUC). We plot this curve by evaluating our newly defined Region-Calibrated score on the Y-axis against the standard PRO score on the X-axis across varying anomaly score thresholds. Next, we detail the computation of the proposed Region-Calibrated score. This score is fundamentally designed to evaluate the average quality of all predicted anomalous regions. Specifically, at a given anomaly score threshold $t$, let $P^t = \{P_1^t, P_2^t, \dots, P_M^t\}$ denote the set of extracted anomalous connected components (regions) from the prediction mask, where $M$ is the total number of predicted regions. The overall Region-Calibrated score at threshold $t$, denoted as $\text{RC}^t$, is computed as the average of the individual calibrated scores across all predicted regions:
\begin{equation}
    \text{RC}^t = \frac{1}{M} \sum_{m=1}^{M} \text{RC}(P_m^t),
\end{equation}
where $\text{RC}(P_m^t)$ is the quality score assigned to the $m^{\text{th}}$ predicted region $P_m^t$. 

To determine this individual quality score, let $T = \{T_1, T_2, \dots, T_N\}$ represent the set of connected ground-truth anomalous regions present in the image, where $N$ is the total number of true defect regions. For a specific predicted region $P_m^t$, if it has absolutely no overlap with any ground-truth anomalous region, its score is penalized to 0, i.e., $\text{RC}(P_m^t) = 0, \text{if } \forall T_n \in T, |P_m^t \cap T_n| = 0$. Conversely, if $P_m^t$ does intersect with the ground truth, we identify its best-matched region, denoted as $T_{n^*}$, which shares the maximum overlapping area: $T_{n^*} = \underset{T_n \in T}{\arg\max} |P_m^t \cap T_n|$. Subsequently, we establish a filtering mechanism to ensure the validity of this prediction. If the overlap between $P_m^t$ and $T_{n^*}$ is deemed negligibly small (i.e., less than a predefined threshold $\tau$, which is empirically set to $0.05 \times |T_{n^*}|$ in our setting), the prediction is also considered a false alarm. Its quality score $\text{RC}(P_m^t)$ is penalized to 0:
\begin{equation}
    \text{RC}(P_m^t) = 0, \quad \text{if } |P_m^t \cap T_{n^*}| < \tau.
\end{equation}
The objective of these conditions is to penalize false positives in normal regions.

For predictions that successfully pass the aforementioned filtering stage, we further evaluate their precision to penalize over-segmentation of anomaly regions. To this end, we compute an excess area ratio, denoted as $\text{E}(P_m^t)$, which quantifies the proportion of the predicted region that redundantly expands beyond the boundaries of the matched ground-truth anomaly $T_{n^*}$:
    $\text{E}(P_m^t) = \frac{|P_m^t| - |P_m^t \cap T_{n^*}|}{|T_{n^*}\vphantom{P_m^t}|}.$
To suppress excessively inflated predictions that dilute localization accuracy, we introduce an inverse decay penalty controlled by an excess tolerance threshold $\theta_{exc}$ (set to 2.0 in our experiments). The final quality score $\text{RC}(P_m^t)$ for a valid predicted region is thus formulated as:
\begin{equation}
    \text{RC}(P_m^t) = 
    \begin{cases} 
      1, & \text{if } \text{E}(P_m^t) \le \theta_{exc} \\
      \frac{\theta_{exc}}{\text{E}(P_m^t)}, & \text{if } \text{E}(P_m^t) > \theta_{exc} 
    \end{cases}
\end{equation}
This mechanism guarantees that a precisely localized prediction within the acceptable tolerance receives a maximum score of 1, whereas predictions that aggressively over-segment the anomalous region are proportionally downgraded.

\end{document}